\documentclass[conference]{IEEEtran}
\IEEEoverridecommandlockouts
\usepackage{balance}
\usepackage{booktabs}
\usepackage{blindtext}
\usepackage{amsmath,amssymb,amsfonts}
\usepackage{algorithmic}
\usepackage{graphicx}
\usepackage{listings}
\usepackage{siunitx}
\usepackage{textcomp}
\usepackage{xcolor}
\usepackage{soul}
\usepackage[backend=biber, natbib=true, style=ieee, doi=true, isbn=false, url=false, citestyle=numeric-comp]{biblatex}

\bibliography{article}

\AtEveryBibitem{%
  \clearfield{note}%
}
\usepackage{hyperref}

\newcommand{\funding}{%
  This work was funded by the EPSRC, grants EP/V052241/1 and EP/S030964/1; and Intel Corporation. TS is also funded by a PhD studentship from the University of Sussex. 
  Compute time was provided through Gauss Centre for Supercomputing application number 21018 and EPSRC (grant number EP/T022205/1) and local GPU hardware was provided by an NVIDIA hardware grant award.
}

\begin{document}

\title{Eventprop training for efficient neuromorphic applications}

\author{%
  \IEEEauthorblockN{%
    Thomas Shoesmith\IEEEauthorrefmark{1},
    James~C.~Knight\IEEEauthorrefmark{1}, Balazs Meszaros\IEEEauthorrefmark{1}, Jonathan Timcheck\IEEEauthorrefmark{2},
    and Thomas Nowotny\IEEEauthorrefmark{1}}
  \IEEEauthorblockA{%
    \IEEEauthorrefmark{1}School of Engineering and Informatics,
                    University of Sussex, Brighton, BN1 9QJ, UK}
    \IEEEauthorrefmark{2}Intel Labs, Intel Corporation, Santa Clara, CA, 95054, USA\\
                    Email: \{t.shoesmith, j.c.knight, b.mszros,t.nowotny\}@sussex.ac.uk, jonathan.timcheck@intel.com   
  \thanks{\funding}
}

\maketitle

\begin{abstract}
Neuromorphic computing can reduce the energy requirements of neural networks and holds the promise to `repatriate' AI workloads back from the cloud to the edge. However, training neural networks on neuromorphic hardware has remained elusive. Here, we instead present a pipeline for training spiking neural networks on GPUs, using the efficient event-driven Eventprop algorithm implemented in mlGeNN, and deploying them on Intel's Loihi 2 neuromorphic chip. Our benchmarking on keyword spotting tasks indicates that there is almost no loss in accuracy between GPU and Loihi~2 implementations and that classifying a sample on Loihi~2 is up to $10\times$ faster and uses $200\times$ less energy than on an NVIDIA Jetson Orin Nano.
\end{abstract}


\section{Introduction}
There has been a growing demand for voice-controlled devices over the last decade, with smart speakers, mobile devices, laptops, and smartwatches all now typically offering voice control. Reliable and accurate Key Word Spotting (KWS) is vital to enable these devices to classify spoken words and phrases to trigger actions. 
Many of these devices operate in an `always on' mode where they listen for a certain command which will wake the device and send the command to the cloud where greater computing resources are available for more complex processing. Although offloading compute to the cloud offers some advantages, it also introduces latency, means that devices can only be used with a stable internet connection and raises privacy concerns.
Reducing the energy requirements of KWS solutions would allow edge devices to perform more processing locally and reduce these concerns. Neuromorphic hardware represents one promising route to achieving this. 

A plethora of neuromorphic systems have been developed by academic groups~\citep{pehle_brainscales-2_2022,gonzalez_spinnaker2_2024,frenkel_reckon_2022,dalgaty_mosaic_2024,dagostino_denram_2024}, startups~\citep{yousefzadeh_seneca_2022,richter_dynap-se2_2024} and larger industrial players~\citep{Merolla2014,davies_advancing_2021}.
These systems use a variety of technologies to implement brain-inspired ``spiking'' neurons which communicate using events which are sparse in space and time. For spatiotemporal tasks such as KWS, using neuromorphic hardware can significantly reduce computation time and energy consumption~\citep{shrestha_efficient_2024}. 
Whilst work in neuromorphic computing has historically been focused on hardware development \cite{Schuman_2022}, key challenges remain in bringing together ideas from computational neuroscience and machine learning for deployment on neuromorphic hardware \cite{Payvand_Neftci_Zenke_2023}. One bottleneck for solving these challenges and developing performant SNN solutions for neuromorphic hardware is the lack of support for end-to-end learning on current systems. We therefore propose a pipeline where SNNs are developed in our user-friendly mlGeNN~\citep{Turner2022,Knight_Nowotny_2023} framework, trained using our efficient GPU implementation of the Eventprop algorithm~\citep{eventProp,nowotny_loss_2024} and then deployed onto the Loihi 2 neuromorphic system. 

In this paper, we demonstrate this pipeline using recurrent SNNs and the Spiking Heidelberg Digits (SHD) and Spiking Speech Commands (SSC) datasets~\citep{Cramer2020}. Both are keyword-spotting tasks where audio recordings of spoken keywords are turned into spike trains using a cochlea model. They have been used extensively as benchmarks for spiking neural network training~\cite{hammouamri_learning_2023,bittar2022surrogate,sun2023learnable,sun2023adaptive,yu2022stsc,nowotny_loss_2024,yao2021temporal,dampfhoffer2022investigating,yin2021accurate,sadovsky2023speech} with current state of the art solutions employing state space models \cite{schone2024scalable} and advanced neuron models \cite{baronig2024advancing,higuchi2024balanced}.
Due to the limited scale of many current neuromorphic systems, much of the previous work on deploying keyword spotting models has focussed on smaller datasets~\citep{blouw_benchmarking_2019,bos_sub-mw_2022}.
However, \citet{patino2024hardware} trained SNN models on SHD with BPTT for deployment on Loihi, \citet{dalgaty_mosaic_2024,dagostino_denram_2024} evaluated SHD classifiers in hardware-aware simulations of future resistive RAM systems and \citet{frenkel_reckon_2022} demonstrated on-chip training on a subset of SHD using e-prop~\citep{Bellec2020}.

Here, we show that running recurrent SNN models on Loihi~2 using our pipeline is around $10\times$ faster and uses about $200\times$ less energy than running the same networks on a Jetson Orin Nano using mlGeNN. Furthermore, the networks retain essentially the same accuracy in spite of the 8-bit quantization of weights and the fixed point arithmetic of states on Loihi.
To the best of our knowledge, this is also the first time that SSC classifiers have been deployed on neuromorphic hardware.

\section{Methods}
All code used for this work is publicly available at \url{https://github.com/neworderofjamie/ml_genn_netx_paper/}, aside from the NxKernel Loihi 2 implementation, which can be made available to members of the Intel Neuromorphic Research Community upon request. 
GeNN, mlGeNN and the mlGeNN to Network Exchange (NetX) converter are available at \url{https://github.com/genn-team/}.
\subsection{Loihi 2}
\begin{figure}
    \centering
    \includegraphics[width=0.85\columnwidth]{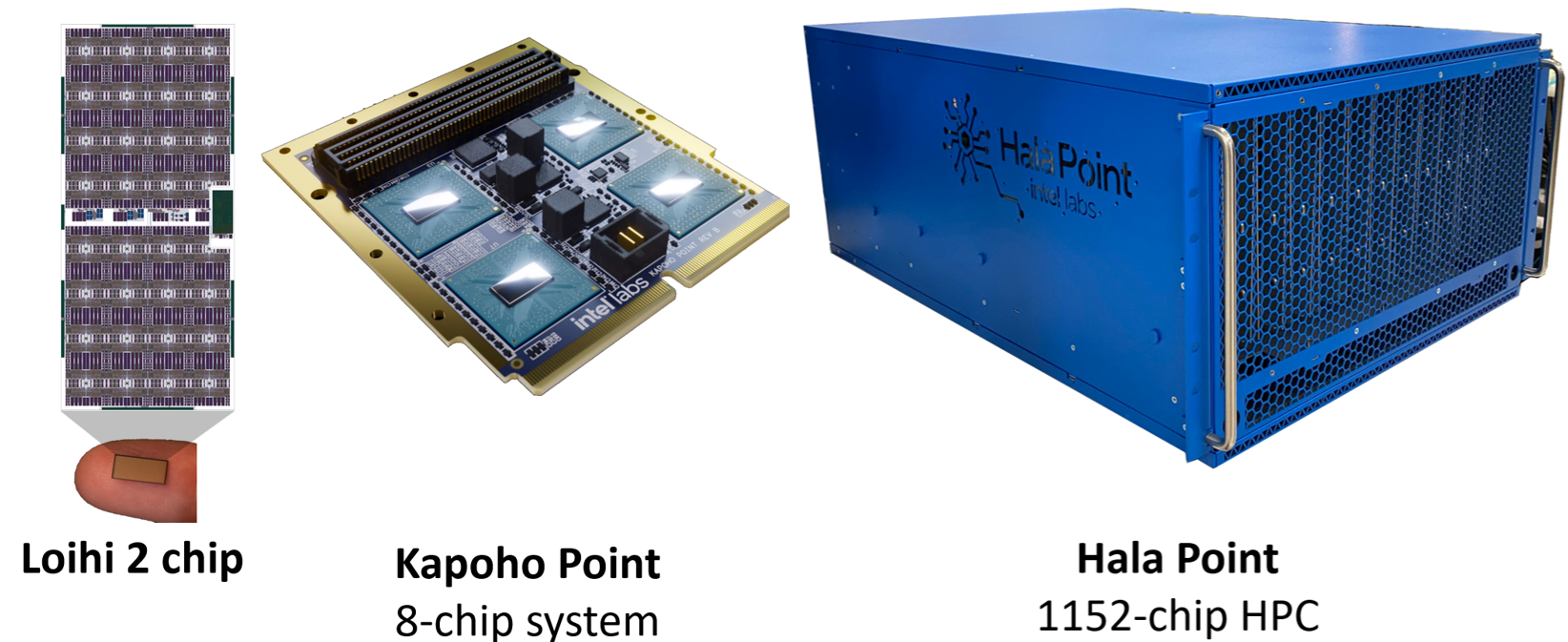}
    \caption{Example Loihi 2 systems, ranging from a 31 mm$^2$ single chip to High Performance Computing (HPC)-scale. Loihi 2 chips can be connected through six asynchronous parallel interfaces, enabling the efficient extension of the Loihi 2 neuromorphic mesh in three dimensions.}
    \label{fig:loihi_2_systems}
\end{figure}

%\href{https://arxiv.org/pdf/2310.03251}{https://arxiv.org/pdf/2310.03251} }
Intel’s latest neuromorphic research chip, Loihi 2, is a fully asynchronous scalable neuromorphic mesh architecture (Figure~\ref{fig:loihi_2_systems}). 
On a Loihi 2 chip, 120 neuromorphic cores execute neuronal dynamics in parallel and communicate with event-based spike messages across a Network-on-Chip. In contrast to more restrictive neuromorphic hardware, Loihi 2 features fully programable neuron models using a rich microcode instruction set and spike messages can optionally carry an integer payload, known as a graded spike. Loihi 2 also has six embedded processor cores which can be used for management and a 10 Gbps Ethernet spike input-output interface.

\subsection{mlGeNN}
mlGeNN~\citep{Turner2022,Knight_Nowotny_2023} is a spike-based ML library built on the GPU-optimised GeNN simulator~\citep{Knight2018,Knight2021}.
Previous publications~\citep{Turner2022,Knight_Nowotny_2023} have described mlGeNN's workflow for converting ANNs to SNNs, its Keras~\citep{chollet2015keras} inspired model description API and `compiler' for training models from scratch using the e-prop~\citep{Bellec2020} learning rule.
We have found that e-prop can train classifiers to competitive performance levels~\citep{Knight_Nowotny_2023} and, as it does not require storing any state between forward and backward passes, its memory requirements do not scale with sequence length.
However, it updates eligibility traces on each connection every timestep meaning that it's computationally expensive and cannot take advantage of the sparse activity of SNNs which GeNN is designed to exploit.

More recently, we have added an Eventprop~\citep{eventProp,nowotny_loss_2024} compiler to mlGeNN.
Eventprop implements a form of event-based Backpropagation Through Time~(BPTT) using exact gradients.
In Eventprop, neuron states only need to be stored between the forward and backward passes at spike times, which massively reduces memory requirements compared to other implementations of BPTT.
Furthermore, the Eventprop backward pass has very similar computational properties to the forward pass of an SNN meaning that it can be implemented very efficiently in GeNN.

\subsection{Datasets}
The Spiking Heidelberg Digits (SHD) and the Spiking [Google] Speech Commands (SSC)~\citep{Cramer2020} datasets are derived from audio recordings and use Lauscher, an artificial cochlea model, to generate spikes over 700 input channels. The SHD dataset comprises \num{10420} samples of studio-recorded spoken digits (0 to 9) in German and English, featuring 12 unique speakers, with two speakers included exclusively in the test set. The SSC dataset is based on the Google Speech Commands~\citep{warden_speech_2018} data, which includes utterances from a larger group of speakers recorded under less controlled conditions. It spans 35 words, providing a broader vocabulary for classification tasks.
\begin{figure}
    \centering
    \includegraphics{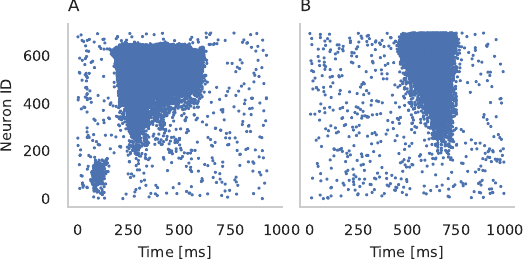}
    \caption{Examples from datasets \textbf{(A)} ``zwei'' from SHD \textbf{(B)} ``left'' from SSC.}
    \label{fig:datasets}
\end{figure}
\begin{table}
\caption{Parameters used to train models of different sizes on SHD and SSC datasets. Regularisation strength is used to initialise both the \lstinline{reg_lambda_upper} and \lstinline{reg_lambda_lower} keyword argument of the \lstinline{EventPropCompiler}.}
  \centering
  \begin{tabular}{r S r S}
    \toprule
        {Dataset} & {Hidden size} &  {Augmentations} & {Regularisation strength}\\
    \midrule
        SHD & 256 & Blend+Shift & 2.5e-10 \\
        SHD & 512 & Blend+Shift & 5e-10 \\
        SHD & 1024 & Blend+Shift & 5e-11 \\
        SSC & 256 & Shift & 2.5e-9\\
        SSC & 512 & Shift & 5e-10 \\
        SSC & 1024 & Shift & 2.5e-10 \\
  \end{tabular}
  \label{tab:training_params}
\end{table}
\subsection{Recurrent Spiking Neural Networks}
\label{sec:rsnn}
\citet{nowotny_loss_2024} showed that, with well-chosen parameters and sufficient augmentation, SNNs with a single recurrently connected hidden layer of Leaky Integrate-and-Fire~(LIF) neurons could be trained to state-of-the-art performance on the Spiking Heidelberg Digits~(SHD) dataset using Eventprop~\citep{eventProp}.
Recurrent architectures of this sort are amongst the best-performing workloads for Loihi~\cite{davies_advancing_2021} which led us to use these architectures for our project.
However, the largest of the models described by \citeauthor{nowotny_loss_2024} have some sophisticated features such as heterogeneous time constants and duplicate inputs with different delays that, while supported by Loihi 2's flexible microcode neuron model, would introduce additional complexity in a first neuromorphic implementation.
We, therefore, train models with 256, 512 and 1024 hidden neurons and the best-performing parameters for models without these features found by \citeauthor{nowotny_loss_2024} (Table~\ref{tab:training_params}).
When using `blend' augmentation, we trained each model for 50 epochs and otherwise for 100 epochs.
In the following subsections, we describe key aspects of the training process in more detail.

\subsubsection{Neuron models}
We use the \lstinline{LeakyIntegrateFire} neuron model and \lstinline{Exponential} synapse model from mlGeNN.
Their dynamics are calculated using an Exponential Euler scheme:
\begin{align}
    I_i[t+1] =& \beta I_i[t] + \sum_j W_{ij} S_j[t] \label{eq:exp_synapse}\\
    V_i[t+1] =& \alpha V_i[t] + (1-\alpha) I_i[t + 1] \label{eq:lif_v}
\end{align}
where $V_i$ is neuron $i$'s membrane voltage, $\alpha = e^\frac{-\text{dt}}{\tau_m}$ controls the neuron's leak ($\tau_m = 20$).
$I_i$ is the neuron's synaptic input current and $\beta = e^\frac{-\text{dt}}{\tau_s}$ controls the timescale of synaptic integration ($\tau_s = 5$). 
The simulation time step $\text{dt} = \SI{1}{\milli\second}$ was used for all experiments.
Spikes are triggered when the membrane voltage crosses the firing threshold $V_\text{th} = 1$ :
\begin{align}
    S[t] =& H\left(V[t] - V_\text{th}\right),
\end{align}
where $H$ denotes the Heaviside function. After spiking, the membrane voltage is reset to 0.
\begin{figure}
    \centering
    \includegraphics{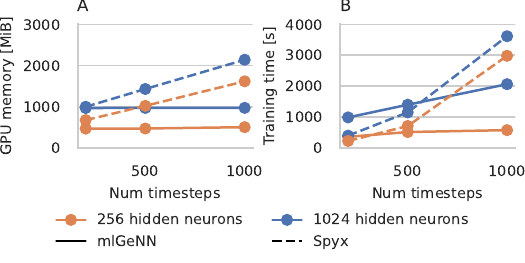}
    \caption{Comparing SHD training cost of Eventprop using mlGeNN against BPTT in Spyx. \textbf{(A)} Peak GPU memory usage, \textbf{(B)} Time to train 100 epochs. All experiments were performed on a workstation with NVIDIA RTX A5000 GPU. All models use batch size 32.}
    \label{fig:spyx_genn}
\end{figure}
\subsubsection{Augmentations}
We used two types of augmentation.
In Shift augmentation, all spikes in each example are shifted along the neuron dimension by an amount drawn from $\text{Unif}[-40,40]$ and rounded to the nearest integer.
In Blend augmentation, the spikes from two randomly chosen examples of the same class are combined into a new spike train by including each spike with \SI{50}{\percent} probability. Before blending, the examples are aligned to their `centre of mass' along both the neuron and time dimensions. 
When blend augmentation was enabled, we added the blended examples to the original training set, effectively doubling its size.
\begin{figure}
    \centering
    \includegraphics{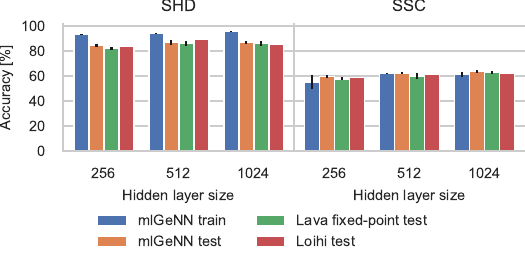}
    \caption{Accuracy of models trained on SHD and SSC and evaluated using mlGeNN using \SI{32}{\bit} floating point as well as using Lava's \lstinline{Loihi2SimCfg} with the ``fixed\_pt'' tag and on Loihi~2 using the NxKernel API. For mlGeNN and Lava results, bar heights represent mean times and error bars standard deviations, all calculated over 5 models trained with a different random seed. On Loihi~2, bar heights represent the accuracy of a single model.}
    \label{fig:accuracy}
\end{figure}

\subsubsection{Learning rate schedule}
To avoid learning failure due to poor initialisation, we `ease in' the learning rate during the first epoch, starting with $\eta = 10^{-6}$ and increasing it by a factor $1.05$ each batch until the final learning rate of $\eta = 10^{-3}$ is achieved. 
When training for more than 50 epochs, we also use a learning rate schedule to reduce the learning rate if the training accuracy plateaus.
This learning rate schedule is driven by a `slow' ($\alpha=0.85$) and `fast' ($\alpha=0.8$) exponentially-weighted moving average~(EWMA) of the training accuracy.
If the fast EWMA ever drops below the slow one and at least $50$ epochs have passed since the last learning rate change, the learning rate is halved.

\begin{table*}
\caption{Spiking Heidelberg Digits classification performance comparison.}
  \centering
  \begin{tabular}{S r r S S S S}
    \toprule
        {Hidden} &  {Weight} & {Hardware} & \multicolumn{4}{c}{Hardware inference cost per sample}\\
        {size} & {precision} & & \multicolumn{2}{c}{Energy}  & {Latency} & {EDP}\\
         & & & {Total [\si{\milli\joule}]} & {Dynamic [\si{\milli\joule}]} & {[\si{\milli\second}]} & {[\si{\micro\joule\times\second}]}\\
    \midrule
        256 & fp32 & Jetson Orin Nano GPU (batch=1)$^\ddagger$ & 82.3 & 23.8 & 25.5 & 2102.0 \\
        512 & fp32 & Jetson Orin Nano GPU (batch=1)$^\ddagger$ & 85.0 & 25.3 & 26.0 & 2214.8 \\
        1024 & fp32 & Jetson Orin Nano GPU (batch=1)$^\ddagger$ & 92.0 & 29.1 & 27.4 & 2522.5\\
        256 & fp32 & Jetson Orin Nano GPU (batch=128)$^\ddagger$ & 3.7 & 1.9 & 103.3 & 384.6 \\
        512 & fp32 & Jetson Orin Nano GPU (batch=128)$^\ddagger$ & 6.5 & 3.7 & 156.4 & 1018.2 \\
        1024 & fp32 & Jetson Orin Nano GPU (batch=64)$^\ddagger$ & 15.5 & 9.7 & 161.2 & 2506.6\\
        256 & int8 & Loihi 2$^\dagger$ & 0.19 & 0.11 & 2.33 & 0.44\\
        512 & int8 & Loihi 2$^\dagger$ & 0.27 & 0.15 & 2.37 & 0.63 \\
        1024 & int8 & Loihi 2$^\dagger$ & 0.50 & 0.31 & 2.56 & 1.29 \\
        \bottomrule \\
  \end{tabular}
  \centering
          \vskip 0.01em 

  \begin{minipage}{0.78\textwidth}
      $^\dagger$Loihi 2 workloads were characterized on an Oheo Gulch system with N3C2-revision Loihi 2 chips running on an unreleased patch of NxCore 2.5.8 and
alpha version of the NxKernel API with on-chip IO unthrottled sequencing of inputs.
 
 $^\ddagger$Jetson workloads were characterized on an NVIDIA Jetson Orin Nano 8GB 15W TDP running Jetpack 5.1.2, GeNN 5.1.0, mlGeNN 2.3.0 and mlGeNN NetX 0.1.0; energy values include CPU\_GPU\_CV and SOC components as reported by jtop.
 
 $^*$Performance results are based on testing as of November 2024 and may not reflect all publicly available security updates; results may vary.
\end{minipage}
  \label{tab:performance}
\end{table*}

\subsection{mlGeNN to NetX conversion} 
\label{sec:ml_genn_netx}
In order to deploy models trained in mlGeNN to Loihi, we have developed an mlGeNN to Network Exchange (NetX) conversion library available at \url{https://github.com/genn-team/ml_genn_netx}.
NetX is an HDF5-based network model exchange format designed for importing models from other frameworks into Lava, an open-source software framework for programming neuromorphic systems available at \url{https://github.com/lava-nc/lava}. 
Due to the hardware constraints of Loihi~2, pretrained weights from mlGeNN needed to be quantised and parameters of the LIF neurons adjusted to match the Loihi implementation. First, we apply Post-Training Quantization~(PTQ) to GeNN’s \SI{32}{\bit} floating point weights to convert them into Loihi’s \SI{8}{\bit} integer weight representation.
While more complex Quantization-Aware Training approaches can be required when quantising weights to below \SI{4}{\bit}, \citet{nagel_white_2021} concluded that simple PTQ of this sort typically results in only minimal reduction in performance when using \SI{8}{\bit} weights.
However, as \citet{McKinstry2018} found, quantization based on the minimum and maximum values is susceptible to outliers reducing the resolution.
Therefore, before quantization to \SI{8}{\bit}, we clipped our weights to within the 99\% percentile. Because incoming connections to each Loihi neuron need to have weights with the same scale, Input to Hidden and Hidden to Hidden weights were scaled together but Hidden to Output weights can be scaled separately. With the scaling of the weights, the hidden neurons' voltage threshold also needs to be scaled by the same factor.

While the dynamics of the Lava ``CUBA-LIF'' model:
\begin{align*}
    V_i[t+1] =& \alpha V_i[t] + I_i[t + 1]
\end{align*}
are \emph{similar} to the mlGeNN dynamics described by equation~\ref{eq:lif_v}, the scaling factor for the current $I$ is missing.
Therefore, as well as converting $\alpha$ (and $\beta$ from equation~\ref{eq:exp_synapse}) to \SI{12}{\bit} fixed point, our mlGeNN to NetX conversion library also pre-multiplies weights $w_{ji}$ by $1 - \alpha$.

\section{Results}
\subsection{Eventprop enables faster training using less memory}
We first assessed the efficiency of our training procedure.
Because Spyx~\citep{heckel_spyx_2024} reportedly has the fastest implementation of Backpropagation Through Time for SNNs, we chose this as our basis of comparison.
For our benchmarking, we used Spyx 0.1.19 and JAX 0.4.26 and modified the SHD classifier model of \citet{heckel_spyx_2024} to include recurrence and LIF neurons with exponential synapses (``RCuBaLIF'') to match the mlGeNN models.
To focus on the most relevant core training, we compared the Spyx model and mlGeNN Eventprop model without augmentations.
We measured the peak memory utilisation of both tools using the  ``nvidia-smi'' command line tool and, when using Spyx, in separate runs with the ``XLA\_PYTHON\_CLIENT\_ALLOCATOR'' environment variable set to ``platform''.
In the Spyx simulations, the entire SHD dataset is uploaded to GPU memory at the start of training, but mlGeNN does not currently support this model of operation.
Therefore, to fairly compare the simulation memory usage and time, we subtract the size of the datasets in memory from the total GPU memory usage for Spyx and exclude the time taken to load the SHD dataset into memory (Spyx) and the time taken to copy examples onto the GPU between batches (mlGeNN).
However, we made sure to include the time taken to generate code and compile it (mlGeNN) or Just-In-Time compile the JAX model (Spyx).

The benchmarking results are illustrated in Figure~\ref{fig:spyx_genn}. Because the memory usage of the Eventprop backward pass scales with the number of events and this remains roughly constant, irrespective of the number of timesteps, the Eventprop memory usage remains essentially constant (Figure~\ref{fig:spyx_genn}A).
However, the memory complexity of the Spyx simulation grows linearly with the number of timesteps as expected for BPTT~\cite{Zenke2021b}.
In all of our experiments with mlGeNN, we allocate enough memory for 750 events per hidden neuron per trial which is conservative, if not excessive, so memory usage in mlGeNN could likely be reduced further. 
For few timesteps, the training time is lower in Spyx but it grows more rapidly with the number of timesteps than in mlGeNN (Figure~\ref{fig:spyx_genn}B). For more than about 500 timesteps, mlGeNN is always faster.

\subsection{Fixed-point conversion has minimal effect on accuracy}
Figure~\ref{fig:accuracy} compares the training accuracies of the models described in section~\ref{sec:rsnn} with the test accuracies obtained using \SI{32}{\bit} floating point weights and quantized \SI{8}{\bit} integer weights obtained using the pipeline described in section~\ref{sec:ml_genn_netx}.
Floating point models were simulated using mlGeNN and quantized models using both Lava's \lstinline{Loihi2SimCfg} with the ``fixed pt'' tag and a Loihi 2 system.
The Loihi 2 networks were implemented using NxKernel, an intermediate-level neuromorphic programming interface that is part of the Lava ecosystem available to Intel Neuromorphic Research Community members. Networks are loaded into the NxKernel implementation through the Lava NetX interface.
Training and floating point test accuracies are comparable to those reported by \citet{nowotny_loss_2024} for these model configurations and similar or better than prior work using `plain' recurrent SNNs with LIF neurons, e.g. 82\% on SHD \citep{zenke2021remarkable} and 57\% on SSC \cite{perez2021neural}. Quantization has no significant impact on test performance, both, in the Lava simulation and on the Loihi~2 system.

\subsection{Loihi~2 delivers low-energy, low-latency inference}
Table~\ref{tab:performance} compares the performance of SHD classification running on Loihi~2.
Results for SSC are essentially the same because the models trained on SSC are the same size, the input is encoded using the same number of neurons and the hidden neuron activity is regularised to the same level. We, therefore, did not repeat this analysis for SSC.
On the Jetson Orin Nano, we measured the runtime of the model using CUDA events at batch size 1, for minimal latency and at several larger batch sizes (32, 64, 128, 256), to find the minimum energy per inference. Following the approach described by \citet{shrestha_efficient_2024}, we measured power usage by adding together the power reported by \lstinline{jtop} on the \lstinline{CPU_GPU_CV} and \lstinline{SOC} rails.
We averaged these values during \SI{20}{\second} of idle time and during the simulation to obtain `static' and `dynamic' power values.
Using these values and the previously calculated simulation times, we calculated the total and dynamic energy per sample and hence the Energy Delay Product of SHD classification.
All of the models simulated here are relatively small so do not fully occupy the Jetson Orin Nano's \num{1024} CUDA core GPU when batch size is 1.
Therefore, latency with batch size 1 is largely limited by the overhead of launching three GPU kernels at each simulation timestep (each kernel incurs around \SI{10}{\micro\second} of latency).
In turn, this means the energy per sample remains relatively constant across model sizes.
However, when using larger batch sizes, this latency is amortised and the energy per sample can be reduced by up to $20\times$ at the expense of higher latency. 

On Loihi, energy and latency were measured on an Oheo Gulch single-chip Loihi 2 system.
A single input sample is loaded into on-chip memory and is subsequently processed repeatedly to achieve accurate, steady-state power and latency measurements; this methodology is also known as the IO-unconstrained mode~\citep{shrestha_efficient_2024}.
Similar to Jetson, the 256, 512, and 1024 hidden neuron workloads are relatively small on Loihi and are readily parallelized across the available neurocores. In fact, all workloads fit in less than half the available neurocores.
This ample parallelization, combined with the network's sparse spiking activity readily handled by the event-based Network-on-Chip mesh, results in the observed slow increase in latency as a function of workload size.

\section{Conclusions and Future Work}
In this work, we have demonstrated a new pipeline for efficiently training recurrent SNNs using Eventprop~\citep{eventProp,nowotny_loss_2024} in mlGeNN~\citep{Turner2022,Knight_Nowotny_2023} and deploying them to the Loihi~2 neuromorphic system.
We have shown that, after quantization, these models can be simulated on Loihi 2~with minimal loss of accuracy and large energy savings compared to conventional processors.
Furthermore, compared to benchmarking of similar RSNN models on the first generation Loihi system~\citep{rao_long_2022}, latency is reduced by approximately $4\times$.

Partly driven by the limitations of BPTT, there has been a trend towards training SNNs with very large timesteps of up to \SI{10}{\milli\second}~\citep{hammouamri_learning_2023,bittar2022surrogate,Yin2021}.
These approaches have shown good performance on benchmark tasks like SHD and SSC because there is, arguably, limited fine temporal information.
However, not only will this approach not suffice for future tasks that depend on fine temporal structure but, the most recent results on SHD and SSC have shown that using precise timing can already lead to higher accuracy~\cite{schone2024scalable,baronig2024advancing}.
In this context, the memory and time savings of Eventprop training that enable working with many small timesteps, combined with the high throughput of Loihi~2 make a particularly promising workflow for neuromorphic edge computing.

Currently, the NetX importer for Lava only supports a limited subset of neuron models and connectivity which restricts the networks that we can convert with the pipeline presented here.
However, in future, we intend to explore training and deploying convolutional and sparsely-connected SNNs as well as SNNs with trained delays and adaptation.
Furthermore, while in this work we have only investigated exporting models to the NetX format for use on Loihi, \citet{pedersen_neuromorphic_2024} have recently developed the Neuromorphic Intermediate Representation~(NIR) which aims to support a wide range of neuromorphic hardware.
It would be trivial to extend our library to export to NIR as well as NetX.
Finally, we have only looked at quantizing synaptic weights down to \SI{8}{\bit} whereas Loihi~2 natively supports smaller data types.
\citet{patino2024hardware} suggest that more advanced Quantization-Aware Training approaches might need to be employed to achieve high accuracy with these levels of quantization.
\balance
\printbibliography

\end{document}